\documentclass[10pt]{article} 
\usepackage[preprint]{tmlr}

\usepackage{amsmath,amsfonts,bm}

\def\eqref#1{equation~\ref{#1}}

\def\1{\bm{1}}

\DeclareMathAlphabet{\mathsfit}{\encodingdefault}{\sfdefault}{m}{sl}
\SetMathAlphabet{\mathsfit}{bold}{\encodingdefault}{\sfdefault}{bx}{n}

\usepackage{booktabs,xcolor}
\usepackage{multirow}
\usepackage{graphicx}
\usepackage{subcaption}
\usepackage{hyperref}
\usepackage{url}
\usepackage{booktabs}

\title{When Do Autoregressive Sequence Models Forecast Physical Wavefields? A Controlled Study on Synthetic Seismograms}

\author{
\name Waleed Esmail\\
\addr Institut f\"ur Kernphysik, Universit\"at M\"unster\\
Wilhelm-Klemm-Stra\ss{}e 9, 48149 M\"unster, Germany
\AND
\name Stuart Russell\\
\addr Institut f\"ur Geophysik, Universit\"at M\"unster\\
Corrensstra\ss{}e 24, 48149 M\"unster, Germany\\
James Cook University, 1 James Cook Drive, Douglas, QLD 4814, Australia
\AND
\name Jana Klinge\\
\addr Institut f\"ur Geophysik, Universit\"at M\"unster\\
Corrensstra\ss{}e 24, 48149 M\"unster, Germany
\AND
\name Alexander Kappes\\
\addr Institut f\"ur Kernphysik, Universit\"at M\"unster\\
Wilhelm-Klemm-Stra\ss{}e 9, 48149 M\"unster, Germany
\AND
\name Christine Thomas\\
\addr Institut f\"ur Geophysik, Universit\"at M\"unster\\
Corrensstra\ss{}e 24, 48149 M\"unster, Germany\\
Geological Survey of Denmark and Greenland, Copenhagen, Denmark
}

\def\month{MM}  
\def\year{YYYY} 
\def\openreview{\url{https://openreview.net/forum?id=XXXX}} 

\begin{document}

\maketitle

\begin{abstract}
Long-horizon autoregressive forecasting of oscillatory physical signals, such as seismograms, gravitational-wave strain, and similar wavefields is limited by error accumulation: as a causal model is fed its own outputs over hundreds of steps, small per-step errors compound into phase drift that pointwise metrics fail to detect. We ask when such rollout stays stable, using synthetic three-component seismograms as a physically structured testbed and the \textsc{SeismoGPT} autoregressive forecaster as the model under study. Through controlled architecture ablations evaluated on free rollout with paired significance tests, we isolate the contribution of each design choice. Multi-token prediction is the dominant stabilizer, accounting for almost the entire improvement over a single-token baseline ($+0.042$ median NCC); a horizon-embedding hybrid prediction head and a cross-horizon STFT-magnitude coherence loss each add a small but consistent further gain. Performance depends sharply on a context-ratio threshold near one, roughly the full P-S interval of observed signal, below which rollout generalization collapses. The dominant residual failure is a polarity inversion: a magnitude-based spectral loss reduces its incidence but, by construction, cannot penalize an inverted segment once it occurs, identifying phase-aware objectives as the natural next step. We frame this as a controlled study of rollout stability on oscillatory wavefields, not a benchmark of forecasting architectures.
\end{abstract}

\section{Introduction}
\label{sec:intro}
Many physical measurement systems produce oscillatory time series that a model must continue, not merely classify: seismograms, gravitational-wave strain, atmospheric pressure records, and biomedical signals such as ECG or EEG all share a wave-like structure in which the quantity of interest is the evolving oscillation itself. Forecasting these signals beyond the span of observed data supports tasks ranging from gap imputation and denoising to source characterization and real-time monitoring. Sequence models, in particular, causal transformers borrowed from language modelling are now a standard tool for this kind of forecasting \citep{nie2023patchtst,das2024timesfm,ansari2024chronos}, and recent work applies them directly to autoregressive continuation of three-component seismograms \citep{esmail2026seismogpt}. At short horizons these models are accurate. The difficulty is the long-horizon rollout regime, where the model is fed its own outputs for hundreds of time steps and must stay faithful to an oscillation it can no longer observe.

In this regime, a specific failure dominates. A causal forecaster trained to predict the next token under teacher forcing sees ground-truth history at training time but its own predictions at inference time, and the resulting train-inference mismatch lets small errors accumulate along the generated sequence \citep{bengio2015scheduled,venkatraman2015dad,lamb2016professor}. For an oscillatory signal, these errors express themselves as phase drift, where each predicted token is slightly mis-timed relative to the underlying oscillation. This mis-timings compound, and after enough steps, the forecast becomes out of phase with the truth. The envelope can still look correct, the model continues to produce a wave of roughly the right frequency content and decay, while the waveform gradually slips out of phase with the target, and in severe cases becomes locally anti-correlated. The phase drift and mismatch we isolate as the dominant residual failure in Section~\ref{sec:exp-setup}. 
Pointwise losses can understate this failure, where small local timing errors may remain modest at each step, while their accumulated effect produces a long-horizon waveform that is visibly out of phase with the target \citep{arora2022exposure,deo2024lossdecomposition}. Phase, more than amplitude, is what these forecasters lose first.

Existing remedies for rollout error largely sit upstream of this problem. Scheduled sampling \citep{bengio2015scheduled} and related curriculum schemes reduce exposure bias by mixing predicted and ground-truth history during training, direct multi-horizon and non-autoregressive forecasters sidestep accumulation by predicting a fixed window in one shot \citep{lim2021tft,BENTAIEB20127067}. None of these targets the boundary between successive predictions as such, instead they treat rollout error as a generic sequence-modelling defect rather than as a spectral phenomenon localized at that boundary. The method we study takes the opposite starting point. \textsc{SeismoGPT} \citep{esmail2026seismogpt} adopts multi-token prediction \citep{qi2020prophetnet,gloeckle2024multitoken,deepseek2024v3}, which exposes the backbone to several future tokens at once during training, and adds two boundary-targeted components: a horizon-embedding hybrid prediction head that gives each horizon its own specialization at the parameter cost of a single shared head, and a multi-resolution Short Time Fourier Transform (STFT) magnitude coherence loss that regularizes the spectral continuity between adjacent predicted horizons. Multi-resolution STFT losses are standard in waveform generation \citep{yamamoto2020parallelwavegan,engel2020ddsp}. Our question is which of these choices actually governs long-horizon stability.

Where \textsc{SeismoGPT} presents the forecasting system, we treat the task purely as a physically structured testbed and ask, through controlled ablations, when such rollout stays stable, isolating which of these design choices governs long-horizon stability, rather than proposing a new system.

Our contributions are scoped to what the controlled ablations support, each effect is established by a paired Wilcoxon signed-rank test with bootstrap confidence intervals over the held-out test events, reported in Section~\ref{sec:exp-setup}. Multi-token prediction is the dominant lever for stabilizing long-horizon rollout on oscillatory signals: almost the entire improvement over a matched single-token baseline is already present once it is enabled. Multi-token prediction itself is established \citep{gloeckle2024multitoken,deepseek2024v3}, our contribution is to quantify, under matched ablations, how strongly it stabilizes long-horizon rollout in this oscillatory physical setting. In addition, two parameter-cheap components of the method each contribute a small but consistent additive gain. First, a horizon-embedding hybrid prediction head that improves on independent per-horizon heads at the parameter cost of a single shared head, in addition, a cross-horizon STFT-magnitude coherence loss. Both effects are small in absolute terms and consistent within a single training run.


The paper is organized as follows. Section~\ref{sec:related} places the work in the context of autoregressive rollout error, tokenized time-series transformers, multi-token prediction, and spectral losses. Section~\ref{sec:method} develops the method. Section~\ref{sec:exp-setup} reports the controlled ablations and the rollout-stability analysis on which the contribution claims rest. Section~\ref{sec:discussion} discusses the mechanism by which the coherence loss dampens phase drift and why it leaves polarity untouched, along with the regime of applicability and the limitations. Section~\ref{sec:conclusion} concludes.

\section{Related Work}
\label{sec:related}

\subsection*{Autoregressive rollout and error accumulation.}
Error accumulation in autoregressive rollout is a long-studied problem. A model trained to predict the next step under teacher forcing sees ground-truth history during training but its own outputs during inference, and that mismatch lets local errors accumulate over a free autoregressive rollout \citep{bengio2015scheduled,lamb2016professor,venkatraman2015dad}. \citep{arora2022exposure} analyze exposure bias from an imitation-learning perspective and show that one-step metrics can fail to reflect error accumulation during generation. The same problem appears in physical forecasting. Autoregressive emulators of spatiotemporal dynamics degrade over long horizons \citep{vlachas2023learning}. In neural PDE solvers, related training-time remedies include exposing the model to rollout-like states during training and predicting multiple future states jointly to improve autoregressive stability \citep{brandstetter2022mppde}. Closer to our metric, \citep{deo2024lossdecomposition} explicitly decompose long-horizon wave-prediction error into phase and amplitude components. What these remedies share is that they act on the training signal in general terms, curricula over predicted history, noise, reweighted error, rather than on where the error lives. Our study asks whether explicitly regularizing spectral continuity across adjacent predicted horizons improves autoregressive rollout.

\subsection*{Tokenized time-series transformers and foundation models.}
A now-standard line of work treats a continuous series as a sequence of tokens. PatchTST \citep{nie2023patchtst} splits the series into sub-window patches, and decoder-only foundation models such as TimesFM \citep{das2024timesfm}, Chronos \citep{ansari2024chronos}, Lag-Llama \citep{rasul2023lagllama}, and Moirai \citep{woo2024moirai} pretrain on large corpora and forecast across domains. These models are primarily developed and evaluated for fixed-horizon forecasting rather than free autoregressive rollout we study, where the model consumes its own output for hundreds of steps. In seismology, deep learning has been most successful at detection, phase picking, and representation learning \citep{zhu2019phasenet,mousavi2020earthquake,liu2024seislm}, autoregressive continuation of full three-component waveforms, our testbed here \citep{esmail2026seismogpt}, remains comparatively underexplored.

\subsection*{Multi-token and multi-horizon prediction.}
Predicting multiple future steps from a single context is a standard strategy in multi-step forecasting. Multi-horizon forecasters such as the Temporal Fusion Transformer \citep{lim2021tft} predict multiple future time steps jointly, while future-$n$-gram objectives such as ProphetNet \citep{qi2020prophetnet} train sequence models to predict several upcoming tokens from the same context; both precede the recent multi-token prediction methods considered here \citep{gloeckle2024multitoken,deepseek2024v3}. A related idea appears in neural PDE solvers as temporal bundling, where the model predicts a block of future states jointly to improve rollout stability and efficiency \citep{brandstetter2022mppde}. Two recent realizations illustrate the design space: \citep{gloeckle2024multitoken} predict multiple future tokens in parallel using independent output heads on a shared trunk, whereas DeepSeek-V3 \citep{deepseek2024v3} predicts additional tokens through sequential Multi-Token Prediction (MTP) modules with shared embedding and output layers. The \textsc{SeismoGPT} head \citep{esmail2026seismogpt} sits between them, a single shared head conditioned on a learned per-horizon embedding. We therefore treat multi-token prediction and existing head variants as prior design motifs.

\subsection*{Spectral and time-frequency losses.}
Multi-resolution STFT losses are a standard ingredient in waveform generation, comparing magnitudes across several time-frequency resolutions to improve fidelity \citep{yamamoto2020parallelwavegan,engel2020ddsp}. Conventionally, the loss scores a completely predicted window. The method's use differs in where it acts, the multi-resolution STFT magnitude loss is applied to concatenations of adjacent predicted horizons, so it scores the boundary between consecutive predictions. The loss family is standard, and the cross-horizon placement was introduced in \textsc{SeismoGPT} \citep{esmail2026seismogpt}. One inherited property matters for our results: a magnitude STFT loss is invariant to absolute phase, so it can dampen boundary discontinuity but cannot, by construction, penalize a polarity-inverted reconstruction \citep{engel2020ddsp}, a limitation we take up in Section~\ref{sec:discussion}. Phase-aware spectral objectives do exist, the speech literature trains directly on wrapped-phase, group-delay, and instantaneous-frequency errors through anti-wrapping losses \citep{ai2023antiwrapping}, and these are the natural basis for the phase-sensitive variant we sketch as future work rather than evaluate here.

\section{Method}
\label{sec:method}
We describe the model as a sequence of design choices, from tokenization to the
training objective. The architecture is the \textsc{SeismoGPT} model
\citep{esmail2026seismogpt}, we restate it only to define the ablations evaluated in Section~\ref{sec:exp-setup}. The two components examined here are the horizon-embedding hybrid prediction head and the cross-horizon STFT-magnitude coherence loss. The contribution of this paper is the matched ablation analysis that isolates their effect on long-horizon rollout, together with the regime and failure-mode analysis reported in Section~\ref{sec:exp-failure}.

\subsection{Problem setup}
\label{sec:setup}
We consider autoregressive forecasting of multi-channel oscillatory time series. Each example is a sequence of $T$ tokens, where each token holds $K$ raw samples across $C$ channels (in this study $K=16$ and $C=3$). Given a context of $T$ tokens, the model predicts the next $H$ tokens jointly during training ($H=4$). At inference, it performs autoregressive rollout, feeding its own predictions back as input. We focus on the rollout regime, in which prediction extends well beyond the training horizon and accumulated errors appear primarily as phase drift and loss of temporal alignment.

\subsection{Tokenization}
\label{sec:tok}
Following \citep{esmail2026seismogpt}, a causal token embedding maps a raw input of shape $(T,C,K)$ to a sequence of $T$ vectors in $\mathbb{R}^{d_{\text{model}}}$. A $1\times1$ convolution first mixes the $C$ channels at each of the $K$ within-token samples, producing features of shape $(d_{\text{model}},K)$, followed by a GELU nonlinearity \citep{Hendrycks:2016qxa}. The $K$ samples are then pooled by concatenating their mean with the final within-token sample, and a linear layer with LayerNorm \citep{ba2016layer} projects the resulting $2d_{\text{model}}$ representation back to $d_{\text{model}}$. The tokenization is identical across the baseline and all ablations, so the comparisons in Section~\ref{sec:exp-setup} isolate the prediction head and objective components rather than the embedding.

\subsection{Backbone}
\label{sec:backbone}
The backbone is a causal transformer encoder with rotary positional embeddings
\citep{su2024roformer}: $d_{\text{model}}=512$, $8$ attention heads, $8$ pre-norm
layers, a GELU feed-forward block with expansion factor $4$, and dropout $0.1$. Rotary
embeddings act on the query and key projections only, and the query, key, value, and
output projections are bias-free. Causal masking ensures that each token attends only to previous tokens. The backbone is deliberately vanilla so that the components of Sections~\ref{sec:head}--\ref{sec:coh} are not confounded with the backbone design applied to the $T$ token embeddings. It produces one hidden state $z_t \in \mathbb{R}^{d_{\text{model}}}$ per position. The architecture follows \citep{esmail2026seismogpt} and is held fixed across all configurations.

\subsection{Multi-token prediction with a horizon-embedding hybrid head}
\label{sec:head}

\begin{figure}[!htbp]
    \centering

    \begin{subfigure}[t]{0.48\linewidth}
        \centering
        \includegraphics[width=0.95\linewidth]{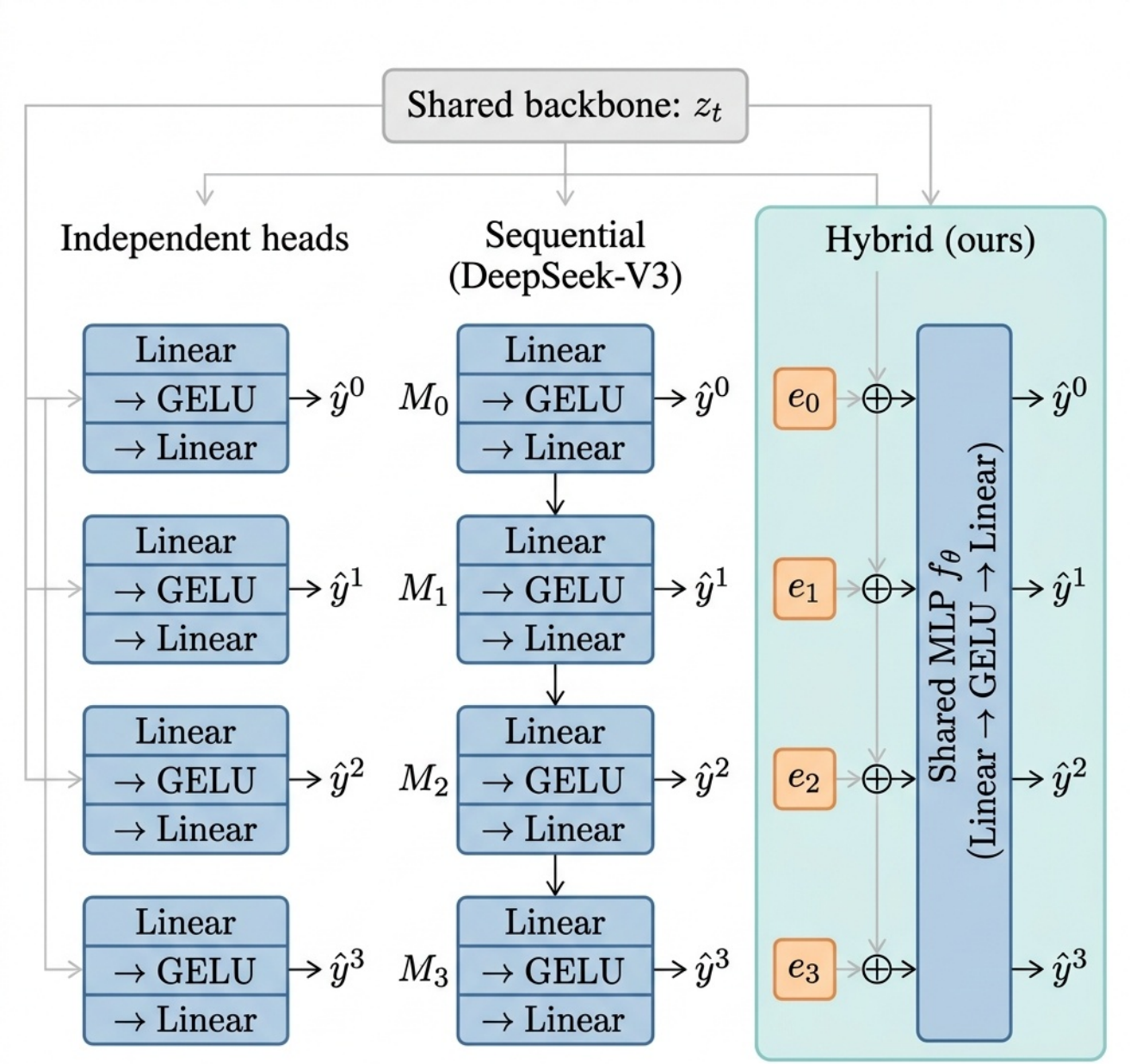}
        \caption{}
        \label{fig:method-a}
    \end{subfigure}
    \begin{subfigure}[t]{0.48\linewidth}
        \centering
        \includegraphics[width=0.95\linewidth]{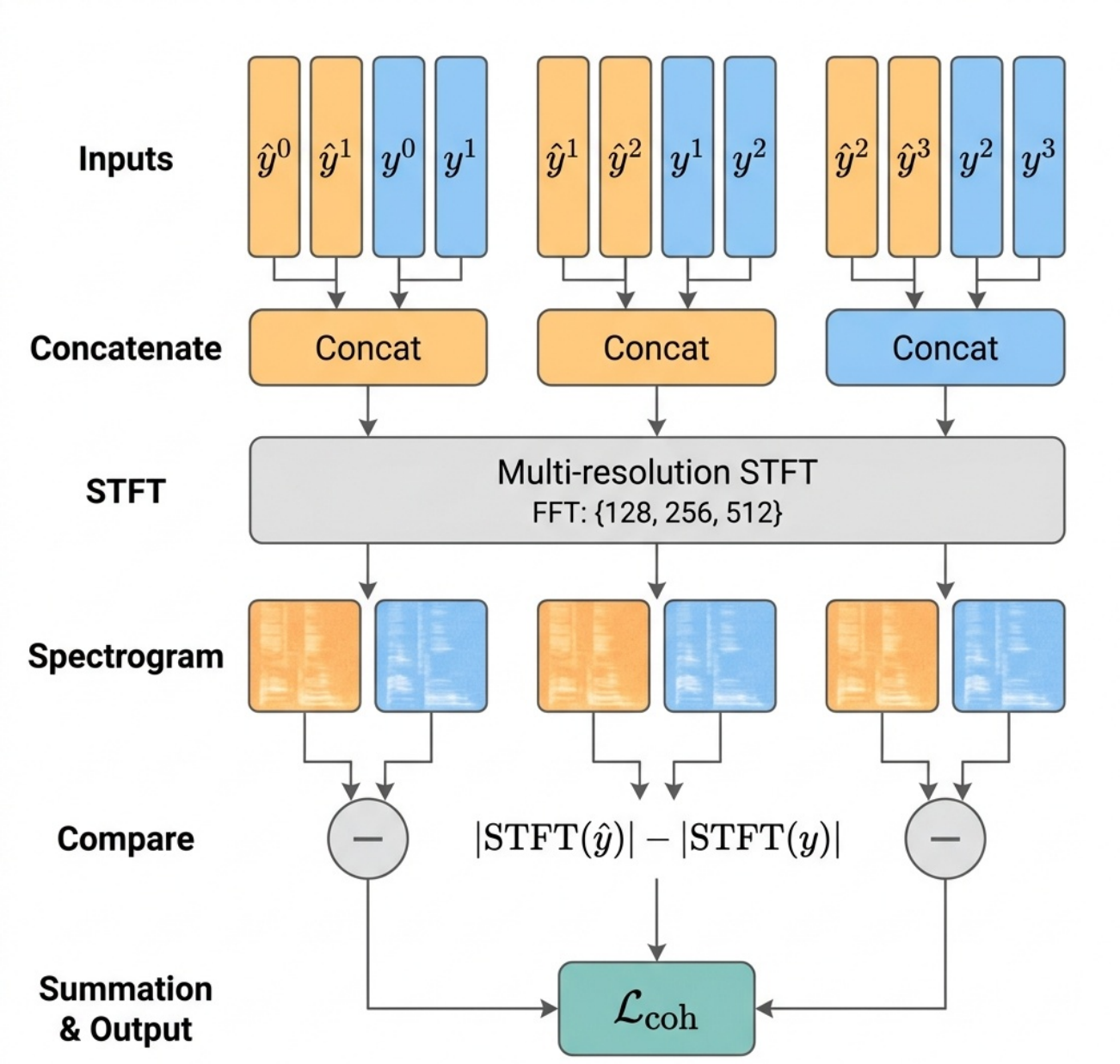}
        \caption{}
        \label{fig:method-b}
    \end{subfigure}

    \caption{The two non-standard components of \textsc{SeismoGPT} \citep{esmail2026seismogpt}. \textbf{(a)} Prediction-head design space. \emph{Independent heads} place $H$ separate MLPs on the backbone in parallel, with no parameter sharing across horizons; the DeepSeek-V3 \emph{sequential} variant chains $H$ modules, each conditioned on the previous module's hidden state; the \emph{hybrid} head applies a single shared MLP $f_\theta$ to $z_t + e_h$, where $e_h$ is a learned per-horizon embedding, giving horizon-specific specialization at the parameter cost of one head. \textbf{(b)} Cross-horizon coherence loss. For each adjacent horizon pair $(h, h{+}1)$, the predicted horizons $\hat{y}^{h},\hat{y}^{h+1}$ and the ground-truth horizons $y^{h},y^{h+1}$ are each concatenated and passed through a multi-resolution STFT (FFT sizes $\{128,256,512\}$); the magnitude spectra are compared by an $L_1$ distance and summed over the three adjacent pairs into $\mathcal{L}_{\text{coh}}$. Because the comparison uses STFT \emph{magnitudes}, it is invariant to absolute phase and cannot penalize the polarity-inverted failure mode (Section~\ref{sec:discussion}).}
    \label{fig:method}
\end{figure}

A standard autoregressive transformer predicts a single next token from each hidden
state $z_t$. \textsc{SeismoGPT} instead predicts the next $H$ tokens jointly (Figure~\ref{fig:method}\subref{fig:method-a}). Each prediction head is a two-layer Multi-Layer Perceptron (MLP) $f_\theta:\mathbb{R}^{d_{\text{model}}} \to \mathbb{R}^{CK}$,

\begin{equation}
f_\theta(z) = W_2\,\phi\!\left(W_1 z + b_1\right) + b_2,
\qquad \phi = \mathrm{GELU},\quad W_1 \in \mathbb{R}^{d_{\text{model}} \times d_{\text{model}}},
\end{equation}

applied at every position $t$ to yield a full-length predicted horizon sequence $\hat{y}^{h} \in \mathbb{R}^{T \times K \times C}$.

Two recent realizations illustrate the design space. \emph{Independent heads} place $H$ separate MLPs on the shared backbone, $\hat{y}^{h}_t = f_{\theta_h}(z_t)$: this multiplies the head parameters by $H$ and shares nothing between horizons. The DeepSeek-V3 \emph{sequential modules} variant conditions the horizon $h{+}1$ on the hidden representation produced for the horizon $h$.

\textsc{SeismoGPT} uses the horizon embedding hybrid head, a single shared MLP $f_\theta$ conditioned on a learned per-horizon embedding $e_h \in \mathbb{R}^{d_{\text{model}}}$, which is added to the hidden state before the head,

\begin{equation}
\hat{y}^{h}_t = f_\theta\!\left(z_t + e_h\right),
\qquad h \in \{0, 1, \dots, H-1\},
\end{equation}

where the $\{e_h\}$ are rows of an embedding table initialized from $\mathcal{N}(0, 0.02^2)$. The shared MLP carries the parameter count of a single independent head, and the $H$ embedding vectors add only $H \cdot d_{\text{model}}$ parameters, so the hybrid head attains per-horizon specialization at the parameter cost of one head rather than $H$. This head is one of the components evaluated in our ablations (Section~\ref{sec:exp-setup}). The multi-horizon training loss is a horizon-weighted sum,

\begin{equation}
\mathcal{L}_{\text{horizon}} = \sum_{h=0}^{H-1} \beta^{h}\, \ell_h,
\qquad \beta = 0.6,
\label{eq:mtp}
\end{equation}

placing greater weight on shorter horizons. The per-horizon term $\ell_h$ is defined
in Section~\ref{sec:objective}; its leading component is a log-cosh distance between
$\hat{y}^{h}$ and the target $y^{h}$, which behaves quadratically near zero and approximately linearly for large residuals.

\subsection{Cross-horizon STFT magnitude coherence loss}
\label{sec:coh}
In autoregressive rollout, drift appears as a growing mismatch in the spectral content between the model's successive predictions, where each horizon may be locally well-predicted, but the spectral content across adjacent horizons drifts from the ground-truth. A loss applied to each horizon in isolation, including the per-horizon STFT auxiliary (Section~\ref{sec:objective}), cannot see this, because it never compares one horizon against the next.

\textsc{SeismoGPT} therefore applies a multi-resolution STFT magnitude loss to the concatenation of adjacent horizon predictions (Figure~\ref{fig:method}\subref{fig:method-b}). The base distance is

\begin{equation}
\mathcal{L}_{\text{STFT}}(x, y)
= \frac{1}{|\mathcal{N}|} \sum_{n \in \mathcal{N}}
\big\langle\, \big|\, |\mathrm{STFT}_n(x)| - |\mathrm{STFT}_n(y)| \,\big|\, \big\rangle,
\label{eq:stft}
\end{equation}

where $\mathcal{N} = \{128, 256, 512\}$ is the set of FFT sizes, a Hann window with hop length $n/4$ is used at each resolution, and $\langle\cdot\rangle$ averages over batch, channel, time, and frequency. The cross-horizon coherence loss evaluates the STFT distance on two-horizon temporal windows formed from adjacent forecast horizons.

\begin{equation}
\mathcal{L}_{\text{coh}} = \sum_{h=0}^{H-2}
\mathcal{L}_{\text{STFT}}\!\left(
[\hat{y}^{h} \,\Vert\, \hat{y}^{h+1}],\;
[y^{h} \,\Vert\, y^{h+1}]
\right),
\label{eq:coh}
\end{equation}

where $\Vert$ denotes temporal concatenation, and each operand has a length of $2TK$ samples per channel. Unlike the per-horizon auxiliary, $\mathcal{L}_{\text{coh}}$ scores adjacent horizons \emph{jointly}, coupling their spectra and penalizing divergence in the joint spectral content as rollout proceeds. Multi-resolution STFT magnitude losses are standard in waveform generation \citep{yamamoto2020parallelwavegan,engel2020ddsp}; the property we evaluate is the cross-horizon placement.

By construction, $\mathcal{L}_{\text{coh}}$ operates on STFT magnitudes, which are invariant to absolute phase. It can therefore dampen spectral drift but cannot detect the polarity-inverted reconstructions we observe among the worst-case rollout events in Section~\ref{sec:exp-setup}, a limitation we return to in Section~\ref{sec:discussion}.

\subsection{Training objective}
\label{sec:objective}
The full training objective combines a horizon-weighted multi-token prediction loss with the cross-horizon coherence term. For each horizon $h$, the per-horizon loss combines the log-cosh base loss with two auxiliary regularizers:

\begin{equation}
\ell_h = \mathcal{L}_{\text{base}}(\hat{y}^{h}, y^{h})
\;+\; \lambda_{\Delta}\, \mathcal{L}_{\Delta}(\hat{y}^{h}, y^{h})
\;+\; \lambda_{\text{stft}}\, \mathcal{L}_{\text{STFT}}(\hat{y}^{h}, y^{h}),
\label{eq:perhorizon}
\end{equation}

where $\mathcal{L}_{\text{base}}(\hat{y}, y) = \big\langle \log\cosh(\hat{y} - y) \big\rangle$ averages over batch, tokens, samples, and channels and $\mathcal{L}_{\Delta} = \mathrm{MSE}\big(\hat{y}^{h}_t - \hat{y}^{h}_{t-1},\,
y^{h}_t - y^{h}_{t-1}\big)$ penalizes mismatch in the first differences across consecutive predicted tokens, and $\mathcal{L}_{\text{STFT}}$ is the per-horizon application of Equation~\ref{eq:stft}. The complete objective is

\begin{equation}
\mathcal{L}
= \underbrace{\sum_{h=0}^{H-1} \beta^{h}\, \ell_h}_{\mathcal{L}_{\text{horizon}}}
\;+\; \lambda_{\text{coh}}\, \mathcal{L}_{\text{coh}},
\qquad
\beta = 0.6,\;\;
\lambda_{\Delta} = \lambda_{\text{stft}} = 0.05,\;\;
\lambda_{\text{coh}} = 0.1.
\label{eq:total}
\end{equation}

The auxiliary terms in Equation~\ref{eq:perhorizon} are included inside the horizon weighting, so their contribution is down-weighted at longer horizons in the same way as the base prediction loss. By contrast, the cross-horizon coherence loss is added outside the per-horizon sum because it is defined on adjacent horizon pairs rather than on individual horizons. The auxiliary losses are treated as part of the training recipe rather than as claimed contributions; their empirical effect is isolated in Section~\ref{sec:exp-setup}.

Training uses AdamW \citep{loshchilov2019decoupled} with a peak learning rate of $10^{-4}$, $1000$ linear warmup steps, cosine annealing with warm restarts and gradient clipping at $1.0$. The same optimizer, schedule, data split, and training protocol are used for the baseline and all ablation configurations. The full experimental details are given in Section~\ref{sec:exp-setup} and Appendix~\ref{app:hparams}.

\section{Experiments}
\label{sec:exp-setup}
We evaluate on synthetic three-component seismic waveforms spanning epicentral distances $10$--$90^\circ$, source depths $5$--$100$\,km, and magnitudes $3$--$7$; the full simulation pipeline is described in \citep{esmail2026seismogpt}. Each trace is tokenized into $K=16$-sample tokens across $C=3$ channels. The model observes a context of $T=320$ tokens and is evaluated on free autoregressive rollouts. All metrics are computed on a disjoint test set using the rollout predictions rather than teacher-forced predictions.

The corpus comprises approximately $7.8$~million simulated events, partitioned into event-disjoint training ($80\%$), validation ($10\%$), and test ($10\%$) splits. Because free autoregressive rollout is expensive to evaluate at scale, all reported rollout metrics are computed on a fixed uniformly random subset of $n=10{,}000$ events drawn from the held-out test split. The same subset is used for every configuration, so all comparisons are paired over identical test events.

For rollout evaluation, the context length is specified relative to the event-specific P-S interval, where P and S denote the primary and secondary seismic phase arrivals, respectively \citep{esmail2026seismogpt}. Let $t_P$ and $t_S$ denote the corresponding arrival times. We define the context ratio $\rho$ by

\begin{equation}
\label{eq:context-ratio}
t_{\mathrm{ctx}} = \rho (t_S - t_P),
\end{equation}

measured from the start of the P-aligned trace. With a token size $K$ and a sampling
rate $f_s$, the number of observed context tokens is

\begin{equation*}
T_{\mathrm{ctx}} =
\left\lfloor
\frac{t_{\mathrm{ctx}} f_s}{K}
\right\rfloor .
\end{equation*}

In our synthetic dataset, $f_s \approx 1.9$ Hz and $K=16$, so one token spans approximately $K/f_s \approx 8.4$s. Thus, the $240$s rollout used in the main evaluation corresponds to approximately $\lfloor 240 f_s/K \rfloor \approx 28$ autoregressive tokens.

We quantify rollout quality with three complementary measures, computed per component (Z, N, E) and channel-averaged over the free autoregressive rollout \citep{esmail2026seismogpt}. Normalized cross-correlation (NCC) measures timing, shape, and phase agreement and is scale-invariant,

\begin{equation}
\mathrm{NCC}(\mathbf{y},\hat{\mathbf{y}})
= \frac{\langle \mathbf{y},\,\hat{\mathbf{y}} \rangle}
       {\|\mathbf{y}\|_2\,\|\hat{\mathbf{y}}\|_2 + \epsilon} \in [-1,1],
\end{equation}

where $\mathbf{y}$ and $\hat{\mathbf{y}}$ are the ground-truth and predicted rollout segments, and $\epsilon$ is a small constant. The signal-to-residual ratio (SRR) adds amplitude sensitivity by treating the prediction residual as noise,

\begin{equation}
\mathrm{SRR}(\mathbf{y},\hat{\mathbf{y}})
= 10\,\log_{10}
\frac{\sum_i y_i^2}{\sum_i (\hat{y}_i - y_i)^2 + \epsilon}.
\end{equation}

The log-spectral PSD error compares the Welch power spectral densities \citep{welch1967use} of prediction and target on a logarithmic scale, after normalizing each density to unit area so that the comparison reflects spectral shape rather than absolute power,

\begin{equation}
\mathcal{E}_{\mathrm{PSD}}
= \frac{1}{N_f} \sum_{k=1}^{N_f}
\big(\log_{10} \tilde{P}_{\hat{y}}(f_k) - \log_{10} \tilde{P}_{y}(f_k)\big)^2,
\qquad
\tilde{P}(f_k) = \frac{P(f_k)}{\sum_{j=1}^{N_f} P(f_j) + \epsilon},
\end{equation}

where $P_{y}(f_k)$ and $P_{\hat{y}}(f_k)$ are the densities at frequency $f_k$ and
$\tilde{P}$ denotes the area-normalized PSD. The three metrics thus separate complementary aspects of forecast quality: NCC captures timing and phase agreement (scale-invariant), SRR captures amplitude fidelity, and $\mathcal{E}_{\mathrm{PSD}}$ captures spectral shape independent of level. Higher NCC and SRR are better; lower $\mathcal{E}_{\mathrm{PSD}}$ is better.

The baseline model is a vanilla causal transformer encoder sharing the same backbone, tokenization, optimizer, and data, differing only in single-token prediction with a log-cosh loss and no auxiliary or coherence terms. We compare four configurations against it. Ablation~A (\emph{no coherence}) is the full model with the cross-horizon coherence loss removed. Ablation~B (\emph{independent heads}) replaces the horizon-embedding hybrid head with $H$ independent per-horizon MLPs. The full model uses the hybrid head and all loss terms. Finally, ablation~C (\emph{MTP only}) adds multi-token prediction with no auxiliary loss and no coherence loss, thereby isolating the effect of multi-token prediction itself. All configurations use the same optimizer, differing only in the prediction head and the active loss terms. Each configuration is trained once due to the computational cost of training at this scale. To reduce evaluation noise, all configurations are evaluated on the same held-out events, and ablation effects are reported as paired differences over matched test examples. We report medians with interquartile ranges and paired bootstrap confidence intervals over test events. This quantifies uncertainty due to finite test-set sampling, but not variability across independent training seeds.

\subsection{Headline results}
\label{sec:exp-headline}
Table~\ref{tab:ablation} reports absolute metrics for all configurations. Table~\ref{tab:significance} reports paired Wilcoxon signed-rank tests with bootstrap $95\%$ confidence intervals for the comparisons that define our analysis. We express effects as paired median $\Delta$NCC because each comparison evaluates two models on the same held-out event. This makes the estimate robust to the heavy left tail of the baseline's per-event NCC distribution (Table~\ref{tab:ablation}). The confidence intervals quantify variability across test events for a single trained model per configuration, however, they do not quantify variability across independent training seeds. We return to this limitation in Section~\ref{sec:discussion}.

Multi-token prediction is the dominant effect. Ablation~C improves the paired median NCC by $+0.042$ over the single-token baseline (95\% CI $[0.040, 0.044]$, $p<0.001$), and the full method improves on the baseline by $+0.047$ (95\% CI $[0.045, 0.049]$, $p<0.001$): almost the entire gain is already present once multi-token prediction is enabled. These results show that it is the dominant mechanism by which the \textsc{SeismoGPT} training recipe transfers to long-horizon waveform forecasting.

\begin{table}[t]
\centering
\caption{Absolute rollout metrics on the test set, per component (Z, N, E) and channel-averaged: NCC and SRR (higher is better), PSD error (lower is better). Values are median (interquartile range); \textbf{bold} marks the best per column. These are descriptive absolute values. Effect sizes and significance for the contribution comparisons are in Table~\ref{tab:significance}.}
\label{tab:ablation}

\newcommand{\val}[2]{#1\,{\scriptsize\textcolor{gray}{(#2)}}}

\begin{tabular}{llccc}
\toprule
Comp. & Config & NCC $\uparrow$ & SRR $\uparrow$ & PSD-norm $\downarrow$ \\
\midrule
\multirow{5}{*}{Z}
 & Baseline    & \val{0.881}{0.391} & \val{7.48}{7.80}  & \val{15.60}{5.40} \\
 & Ablation A  & \val{0.937}{0.164} & \val{10.07}{6.00} & \val{13.98}{5.32} \\
 & Ablation B  & \val{0.930}{0.156} & \val{9.63}{5.54}  & \val{13.02}{4.93} \\
 & Ablation C  & \val{0.945}{0.138} & \val{10.64}{5.84} & \val{15.99}{6.94} \\
 & \textsc{SeismoGPT} & \val{\textbf{0.948}}{0.119} & \val{\textbf{10.89}}{5.75} & \val{\textbf{12.65}}{4.64} \\
\addlinespace
\multirow{5}{*}{N}
 & Baseline    & \val{0.906}{0.288} & \val{8.39}{8.01}  & \val{14.71}{3.66} \\
 & Ablation A  & \val{0.959}{0.120} & \val{11.79}{7.04} & \val{12.33}{3.55} \\
 & Ablation B  & \val{0.952}{0.119} & \val{11.17}{6.43} & \val{12.04}{3.60} \\
 & Ablation C  & \val{0.963}{0.101} & \val{12.24}{6.78} & \val{14.06}{4.26} \\
 & \textsc{SeismoGPT} & \val{\textbf{0.966}}{0.082} & \val{\textbf{12.67}}{6.34} & \val{\textbf{11.36}}{3.13} \\
\addlinespace
\multirow{5}{*}{E}
 & Baseline    & \val{0.894}{0.307} & \val{7.93}{8.05}  & \val{14.80}{4.19} \\
 & Ablation A  & \val{0.956}{0.124} & \val{11.73}{7.20} & \val{12.41}{4.06} \\
 & Ablation B  & \val{0.949}{0.129} & \val{11.00}{6.56} & \val{12.09}{4.01} \\
 & Ablation C  & \val{0.961}{0.107} & \val{12.27}{7.11} & \val{13.98}{5.29} \\
 & \textsc{SeismoGPT} & \val{\textbf{0.965}}{0.088} & \val{\textbf{12.62}}{6.56} & \val{\textbf{11.56}}{3.87} \\
\midrule
\multirow{5}{*}{\textbf{Mean}}
 & Baseline    & \val{0.886}{0.352} & \val{8.20}{7.54}  & \val{15.15}{4.26} \\
 & Ablation A  & \val{0.949}{0.140} & \val{11.48}{6.04} & \val{13.13}{3.95} \\
 & Ablation B  & \val{0.942}{0.144} & \val{10.88}{5.45} & \val{12.57}{3.94} \\
 & Ablation C  & \val{0.954}{0.121} & \val{12.01}{5.86} & \val{14.89}{5.30} \\
 & \textsc{SeismoGPT} & \val{\textbf{0.959}}{0.099} & \val{\textbf{12.41}}{5.41} & \val{\textbf{12.07}}{3.47} \\
\bottomrule
\end{tabular}


\end{table}

\begin{table}[t]
\centering
\caption{Paired Wilcoxon signed-rank tests over the test events, with bootstrap $95\%$ confidence intervals on the median per-event $\Delta$NCC. These are the effect sizes our contribution claims cite. Each comparison isolates one component: MTP (C vs.\ baseline), the hybrid head (Full vs.\ B), and the coherence loss (Full vs \ A).}
\label{tab:significance}
\begin{tabular}{lrrrrr}
\toprule
Comparison & $n$ & $W$ & $p$ & Median $\Delta$NCC & 95\% CI \\
\midrule
Full vs Ablation A & 10000 & 17029772 & $<0.001$ & $+0.006$ & $[0.006, 0.007]$ \\
Full vs Ablation B & 10000 & 14946375 & $<0.001$ & $+0.011$ & $[0.010, 0.011]$ \\
Full vs Ablation C & 10000 & 21601752 & $<0.001$ & $+0.002$ & $[0.002, 0.003]$ \\
Ablation C vs Baseline & 10000 & 5824239 & $<0.001$ & $+0.042$ & $[0.040, 0.044]$ \\
Full vs Baseline & 10000 & 5255807 & $<0.001$ & $+0.047$ & $[0.045, 0.049]$ \\
\bottomrule
\end{tabular}

\end{table}

Within the multi-token setting, the two studied components provide small but consistent additional gains. The horizon-embedding hybrid head improves on
independent per-horizon heads by $+0.011$ NCC (95\% CI $[0.010, 0.011]$, $p<0.001$)
and the cross-horizon coherence loss improves on the otherwise identical no-coherence model by $+0.006$ NCC (95\% CI $[0.006, 0.007]$, $p<0.001$). The auxiliaries and the coherence loss together add only $+0.002$ NCC over multi-token prediction alone (Full vs.\ Ablation~C; 95\% CI $[0.002, 0.003]$, $p<0.001$). In absolute terms (Table~\ref{tab:ablation}), the auxiliaries on their own do not help NCC, adding them to the MTP-only model lowers median NCC from $0.954$ (Ablation~C) to $0.9494$ (Ablation~A, which differs from C only by the auxiliaries), and a net gain appears only once the coherence loss is also present ($0.959$, full). For NCC, then, the auxiliaries are justified empirically only when combined with the coherence loss.

Because n = 10,000 paired events makes even very small differences statistically significant, we also report how consistent each effect is at the level of individual events. The full model attains higher per-event NCC than the single-token baseline on 86.8\% of the test events, than Ablation B on 70.1\%, than Ablation A on 64.2\%, and than Ablation C on 56.1\%. The MTP effect is therefore both large and nearly universal across events, whereas the component effects are small but directionally consistent across the majority of the test population.

\subsection*{Failure tail}
Because ``stabilizing rollout'' is a claim about failures rather than about the median event, Table~\ref{tab:tail} reports the left tail of the per-event NCC distribution for each configuration. Multi-token prediction collapses the baseline's heavy failure tail: the probability of a badly degraded rollout, $P(\mathrm{NCC}<0.5)$, drops from $20.2\%$ (baseline) to $8.8\%$ (Ablation~C), and the 5th-percentile NCC rises from $0.05$ to $0.32$. The two studied components act primarily in this tail as well: the full model attains the lowest $P(\mathrm{NCC}<0.9)$ ($28.3\%$), the lowest $P(\mathrm{NCC}<0.5)$ ($7.9\%$), and the highest 5th-percentile NCC ($0.34$) of all configurations, consistent with the interpretation that the head and the coherence loss slow degradation on hard events rather than improving typical-case accuracy. In the extreme tail the picture is mixed, ablation~C is marginally best on $P(\mathrm{NCC}<0)$ ($1.2\%$ vs.\ $1.5\%$ for the full model) and on the 1st percentile, so the components improve the bulk of the failure tail but do not reduce the rarest catastrophic failures. This tail view also contextualizes the small median effects of Table~\ref{tab:significance}. Most events are forecast accurately by every multi-token configuration, so population medians can move only slightly.

\begin{table}[t]
\centering
\caption{Failure-tail statistics of channel-averaged per-event NCC on the free-running $240$\,s rollout ($n=10{,}000$ paired events). Lower failure probabilities and higher percentiles are better.}
\label{tab:tail}
\begin{tabular}{lrrrrr}
\toprule
Config & $P(\mathrm{NCC}<0.9)$ & $P(\mathrm{NCC}<0.5)$ & $P(\mathrm{NCC}<0)$ & 5th pctl & 1st pctl \\
\midrule
Baseline & 53.2\% & 20.24\% & 3.98\% & 0.051 & -0.238 \\
Ablation C & 31.3\% & 8.78\% & 1.24\% & 0.319 & -0.047 \\
Ablation A & 33.9\% & 10.03\% & 1.79\% & 0.251 & -0.114 \\
Ablation B & 35.1\% & 9.88\% & 1.64\% & 0.270 & -0.085 \\
Full & 28.3\% & 7.86\% & 1.47\% & 0.337 & -0.075 \\
\bottomrule
\end{tabular}

\end{table}

\subsection*{Spectral fidelity} 
The normalized PSD-shape error gives a complementary picture. Unlike NCC and SRR, which are dominated by the effect of multi-token prediction, PSD error improves monotonically as the additional loss terms are introduced. The channel-averaged PSD error decreases from $15.1$ for the baseline to $14.9$ for Ablation~C (MTP only), then to $13.1$ for Ablation~A after adding the auxiliary terms, and finally to $12.1$ for the full model. The full model is also the best on each individual component.

This separates the role of the auxiliary losses from the role of MTP. MTP is the main driver of NCC and SRR, but the auxiliary losses substantially improve spectral shape, reducing PSD error from $14.9$ to $13.1$. In isolation, these auxiliary losses slightly reduce NCC and SRR relative to MTP only, but their spectral benefit is retained when the coherence loss is added. The full model therefore combines the strongest correlation metrics with the best spectral shape: SRR increases from $8.20$\,dB for the baseline to $12.01$\,dB for MTP only and $12.41$\,dB for the full model, while PSD error falls to $12.1$. We therefore retain the auxiliary losses on spectral grounds rather than because they improve NCC by themselves.

\subsection*{Rollout stability}
\label{sec:exp-rollout}

\begin{figure}[!bht]
    \centering
    \includegraphics[width=0.9\linewidth]{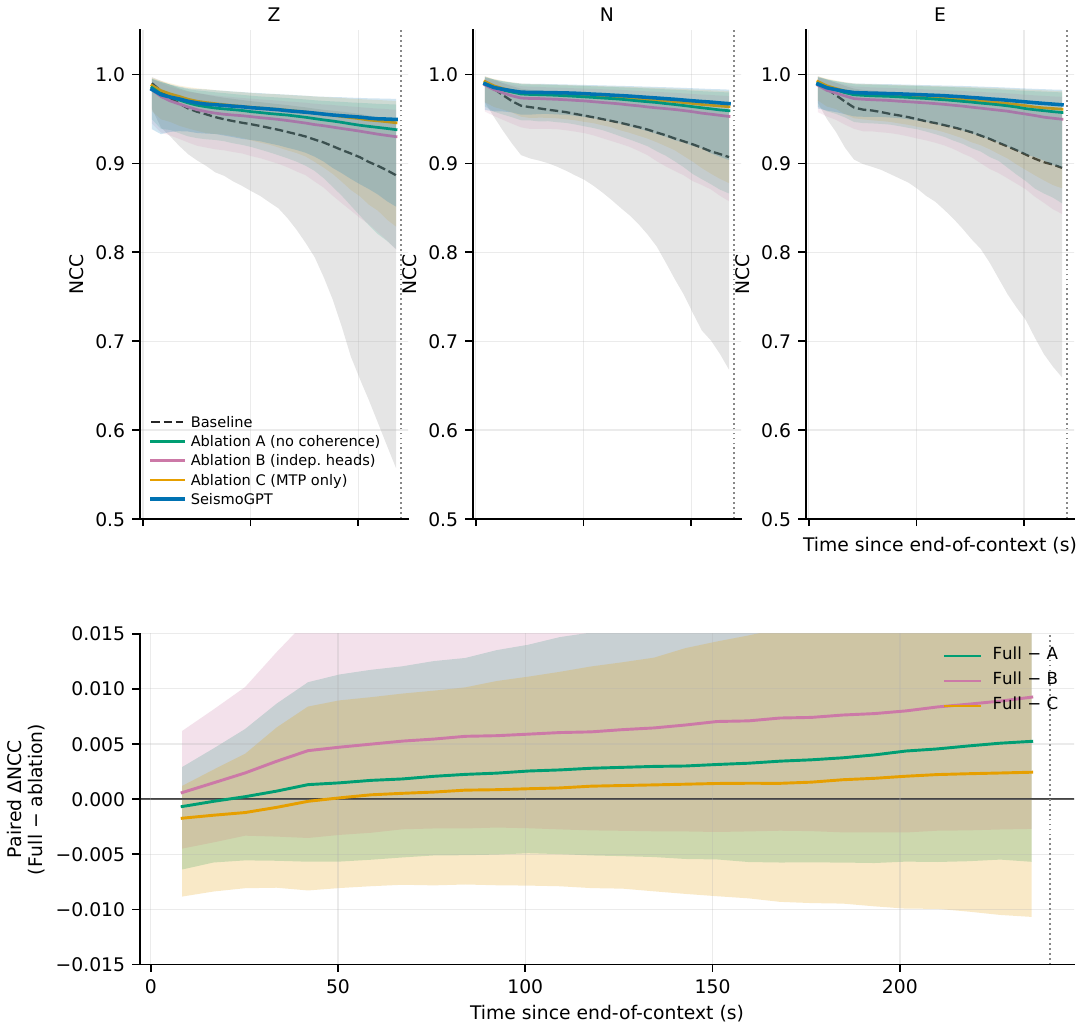}
    \caption{Rollout stability across the autoregressive continuation. \textbf{Top:} NCC per component (Z, N, E) versus time since the end of the context window for the baseline, ablations, and full model. \textbf{Bottom:} paired $\Delta$NCC of the full model relative to each ablation (Full~$-$~A, Full~$-$~B, Full~$-$~C). Curves show medians over test events and shaded bands show interquartile ranges. The gaps are small but become mostly positive as the rollout extends, providing visual support for the modest contribution claims in Table~\ref{tab:significance}.}
    \label{fig:rollout}
\end{figure}

Figure~\ref{fig:rollout} resolves the aggregate numbers across rollout time. The top row plots NCC per component against time since the end of the context window: all multi-token configurations maintain high median NCC with small separation among multi-token variants, while the single-token baseline degrades faster as the rollout lengthens. The bottom row plots the paired $\Delta$NCC between the full model and each ablation over the same axis. The differences are small within roughly $\pm0.01$ NCC, consistent with Table~\ref{tab:significance}, but the full model remains at or above each ablation across most of the rollout. This panel is the temporal counterpart of the aggregate paired tests: it is consistent with the components slowing degradation as the forecast extends, not by changing one-step accuracy. Consistent with this, a per-horizon breakdown (Appendix~\ref{app:horizon}) finds the configurations essentially indistinguishable at every individual horizon (all gaps $<0.001$ NCC); the contributions manifest in rollout dynamics, not per-step accuracy, which is why we omit a per-horizon column from Table~\ref{tab:ablation}.

\subsection{Qualitative forecasts}
\label{sec:exp-qualitative}
\begin{figure}[!bht]
    \centering
    \includegraphics[width=0.8\linewidth]{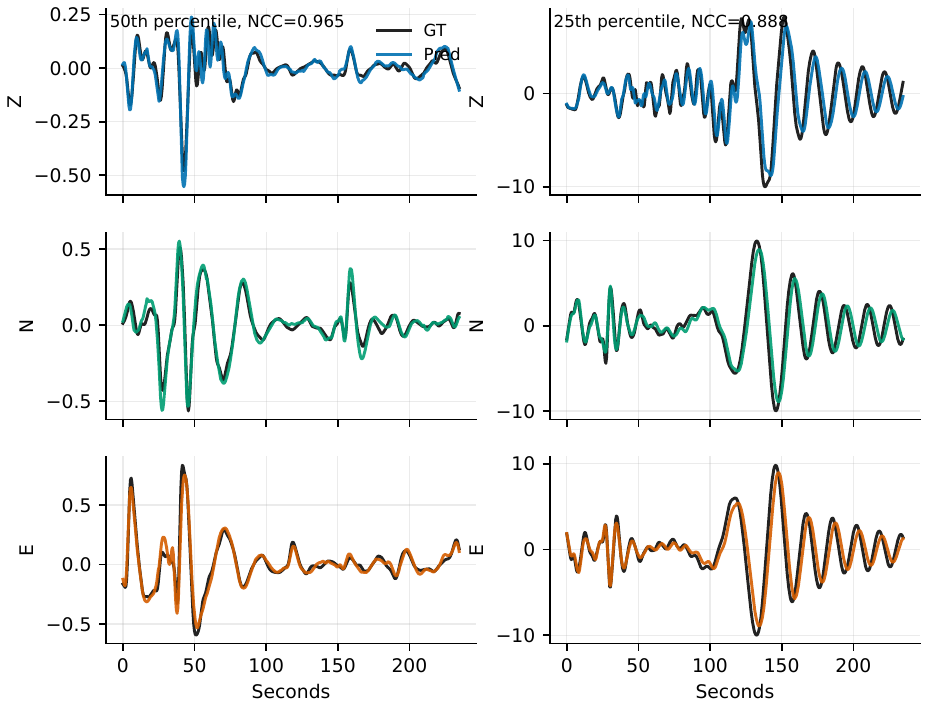}
    \caption{Full-model rollout versus ground truth for a median event (50th percentile, NCC $=0.965$, left) and a lower-quartile event (25th percentile, NCC $=0.888$, right), across all three components. Ground truth is shown in black and predictions in component-specific colors.}
    \label{fig:qualitative}
\end{figure}

Figure~\ref{fig:qualitative} shows the full model's free-running rollout against ground truth for a median event (50th percentile, NCC $=0.965$) and a lower-quartile event (25th percentile, NCC $=0.888$). The two events were selected at these percentile ranks during an earlier evaluation pass; under the final evaluation run the population median NCC is $0.959$, so the plotted median event sits marginally above the final median. We state the percentile ranks to make the selection criterion explicit. The median case tracks both phase and amplitude across all three components. Even the lower-quartile case remains well-correlated, with errors concentrated in the later part of the rollout. The failures that dominate the lower tail of the distribution are qualitatively different and are analyzed below.

\subsection{Regime of applicability}
\label{sec:exp-context}
\begin{figure}[!bht]
    \centering
    \includegraphics[width=0.7\linewidth]{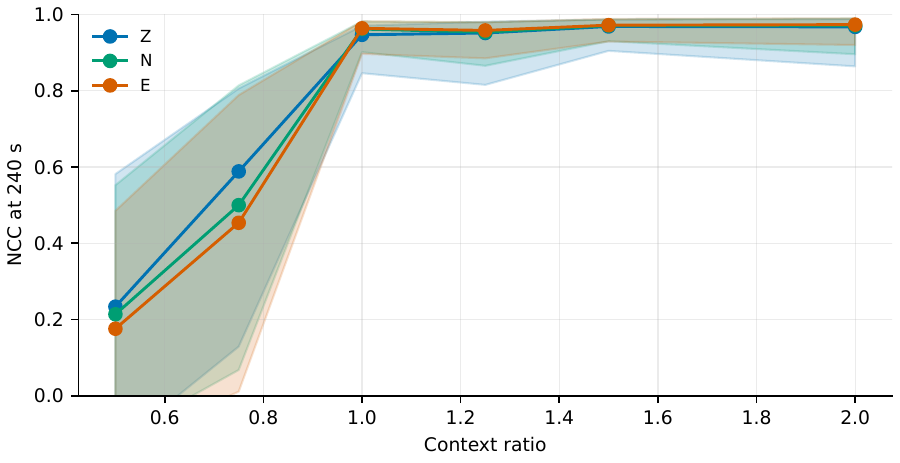}
    \caption{NCC at a fixed $240$\,s rollout horizon as a function of the context ratio $\rho$, where $\rho$ sets the observed context length relative to the event-specific P-S interval. Shading shows the interquartile range. Performance is high and nearly flat for $\rho \geq 1$ and degrades sharply below this threshold.}
    \label{fig:context}
\end{figure}

Figure~\ref{fig:context} characterizes the regime in which the method is effective by sweeping the context ratio (Equation~\ref{eq:context-ratio}), and measuring NCC at a fixed $240$\,s rollout horizon. Performance shows a sharp threshold near $\rho=1$. For $\rho \geq 1$, the median NCC remains high and nearly flat across components, whereas for $\rho < 1$ it degrades sharply and exhibits a substantially larger spread. This threshold indicates that the model requires approximately the full P-S interval before reliable long-horizon rollout.

The threshold has a natural physical reading, with $\rho \geq 1$ the context containing both the P and the S arrivals, so the observed waveform implicitly constrains the source–receiver geometry and the relative timing and polarization of the dominant phases. However, with $\rho < 1$ the S arrival, which carries most of the energy the model must continue has not yet been observed, and the continuation problem becomes underdetermined. We state this as an interpretation consistent with the sweep, not as an established mechanism.

\subsection{Failure analysis}
\label{sec:exp-failure}
A recurring residual failure mode is phase drift and mismatch rather than amplitude decay. In the worst-case events (Appendix~\ref{app:failures}, Figure~\ref{fig:failure}), the predicted late high-amplitude segments anti-correlate with the ground truth, driving NCC negative (down to $\approx -0.56$). The model continues to emit oscillatory waveforms with roughly plausible frequency content, but they become phase-shifted or sign-flipped over high-energy portions of the rollout. Because the cross-horizon coherence loss operates on STFT magnitudes, which discard phase, it cannot, by construction, penalize these polarity-flipped reconstructions. This is the mechanistic limitation we return to in Section~\ref{sec:discussion}.

\subsubsection{Quantifying phase drift and polarity inversion.} 
\begin{figure}[!bht]
    \centering
    \includegraphics[width=\linewidth]{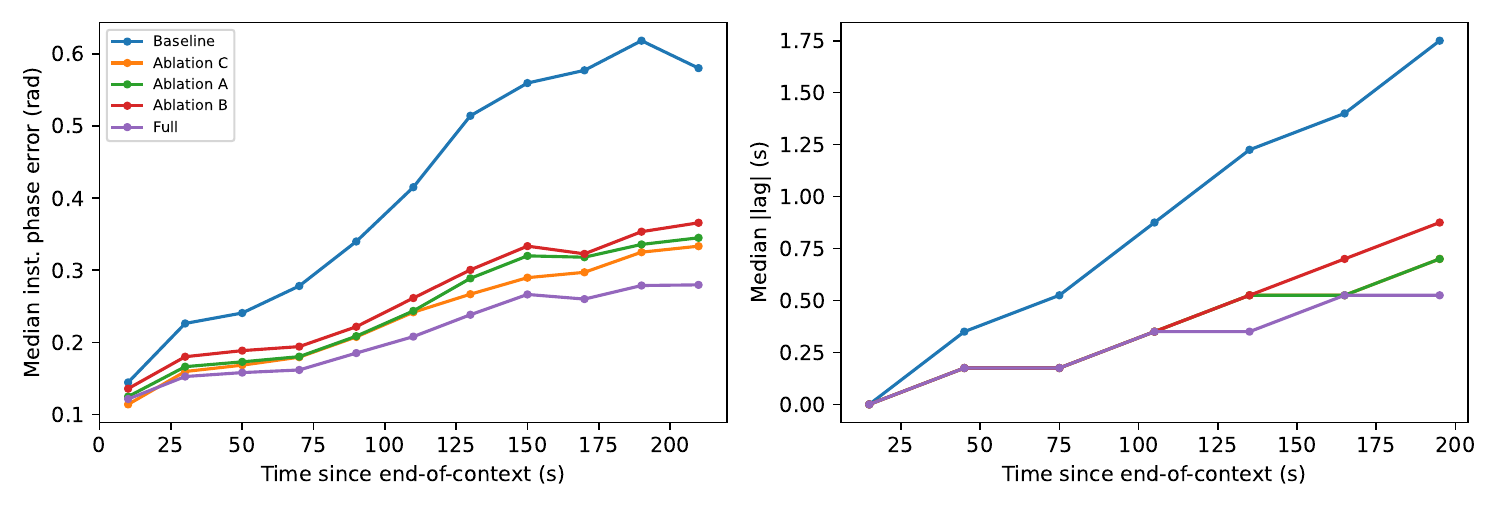} 
    \caption{Direct measurement of phase drift over the autoregressive rollout. \textbf{Left:} median energy-weighted instantaneous phase error (radians), computed from the analytic signal in $20$\,s windows, versus time since end-of-context, for the baseline, the three ablations, and the full model. \textbf{Right:} median absolute best-alignment lag (seconds) from windowed cross-correlation ($30$\,s windows, $\pm10$\,s search). The single-token baseline accumulates phase error and lag steadily as the rollout extends, whereas all multi-token configurations remain nearly flat, showing that multi-token prediction stabilizes rollout specifically by suppressing phase drift.}
    \label{fig:phase-drift}
\end{figure}

For each event, we compute the energy-weighted instantaneous phase error between prediction and ground truth via the analytic signal, and the best-alignment lag of the prediction within sliding windows, both as functions of time since the end of the context (Figure~\ref{fig:phase-drift}). For the single-token baseline the median instantaneous phase error grows monotonically from $0.14$\,rad in the first $20$\,s window to $0.58$\,rad in the final window (peaking at $0.62$\,rad), whereas all multi-token configurations remain below $0.37$\,rad throughout, with the full model lowest at $0.28$\,rad; the windowed lag shows the same separation, reaching a median of $1.75$\,s for the baseline against $0.53$--$0.88$\,s for the multi-token configurations. This confirms that the improvement measured by NCC is specifically a phase-stability effect rather than an amplitude effect. We also quantify the polarity failure mode as a population statistic. An event is counted as polarity-inverted if, over the final half of the rollout, the sign-flipped prediction correlates with the ground truth better than the prediction itself by a margin of $0.1$ NCC on at least one component. MTP is again the dominant lever, reducing the flip rate from $21.3\%$ (baseline) to $9.5\%$ (Ablation~C). Contrary to the simplest reading of the magnitude-invariance argument, the coherence loss is associated with a further reduction: the flip rate is $7.8\%$ for the full model versus $9.8\%$ for Ablation~A, and a paired exact test finds this difference significant ($p<0.001$). We interpret this carefully in Section~\ref{sec:discussion}; magnitude invariance implies that the loss cannot penalize an already-inverted segment, not that it cannot reduce the incidence of trajectories that drift into inversion. A $7.8\%$ flip rate nonetheless remains the dominant residual failure, motivating the phase-aware objectives discussed there.

\section{Discussion}
\label{sec:discussion}
In the standard objective, each forecast horizon is evaluated separately against its corresponding target. The cross-horizon coherence loss instead evaluates adjacent horizons jointly through the multi-resolution STFT magnitude of their temporal concatenation. It is therefore sensitive to spectral structure across horizon boundaries, rather than only to the spectral content of each horizon separately. This is intended to address one failure mode of long-horizon autoregressive rollout, where errors accumulate. The predicted time-frequency energy can drift away from the ground-truth distribution even when individual short-horizon predictions remain accurate. Because the loss is defined on STFT magnitudes, it can only act on the component of rollout drift that appears as a mismatch in time-frequency magnitude.

The same construction also limits what the loss can correct. Because it compares STFT magnitudes, it discards phase information. In particular, $|\mathrm{STFT}(-x)| = |\mathrm{STFT}(x)|$ for a global sign flip. Thus, a prediction with plausible spectral magnitude but incorrect phase can receive a small coherence penalty while scoring poorly under NCC. This matches the failure analysis in Section~\ref{sec:exp-failure}, where the lowest-NCC events are not simple amplitude-decay failures, but phase-drift failures in which late high-energy segments can become locally anti-correlated or sign-flipped. This helps explain why the coherence-loss effect is small but consistent ($+0.006$ NCC, Section~\ref{sec:exp-headline}), where only the magnitude-visible component of rollout drift is directly targeted. The polarity-flip analysis of Section~\ref{sec:exp-failure} sharpens this argument in an instructive way. The magnitude-only construction implies the loss assigns no penalty to an already sign-inverted segment; it does not imply the loss cannot affect how often rollouts drift into inversion. Empirically we observe the latter: adding the coherence loss reduces the population flip rate from $9.8\%$ to $7.8\%$ ($p<0.001$), plausibly because damping spectral discontinuities at horizon boundaries reduces the accumulated drift that precedes an inversion, while the phase-drift curves (Figure~\ref{fig:phase-drift}) show that the multi-token recipe as a whole acts specifically by suppressing accumulated phase error. Because this is a single training run, we offer this incidence effect as an observation consistent with the boundary-damping mechanism rather than as an established causal claim. What remains true by construction is that once an inversion has occurred, the coherence loss exerts no corrective pressure on it, which is why a phase-aware objective remains the natural next step. A phase-aware extension, such as a complex-STFT term, group-delay term, or explicit sign/correlation penalty, in the spirit of anti-wrapping objectives used in waveform synthesis \citep{ai2023antiwrapping}, is the natural next step. 

The main empirical conclusion is that multi-token prediction stabilizes long-horizon autoregressive rollout for this oscillatory physical forecasting task, and the horizon-embedding hybrid head and cross-horizon coherence loss add small but consistent gains on top of it. The context-ratio sweep (Section~\ref{sec:exp-context}) makes the regime of applicability explicit, performance remains high for $\rho \geq 1$ and degrades sharply below this threshold. Since $\rho$ is defined relative to the event-specific P-S interval, this means that a reliable $240$\,s rollout requires approximately the full P-S interval as context.

The limitations are also specific. First, the evaluation uses a single synthetic dataset, real recordings and additional datasets are required before the effects can be treated as general. Second, this is a controlled ablation study within one architectural class, a causal transformer encoder. We do not compare against external sequence-model baselines, so the results show that the studied components help within this GPT-style autoregressive model, not that this model class is preferable to alternative forecasters. Comparisons against state-space models such as Mamba \citep{gu2023mamba}, direct multi-horizon forecasters, and non-autoregressive sequence models are natural next experiments. Third, each configuration is trained once under a fixed seed. The reported confidence intervals quantify variability across held-out test events, not variability across training initializations or data order. The small component effects ($+0.011$ for the head and $+0.006$ for the coherence loss) should therefore be read as within-run effects and confirmed across seeds before being treated as fully established. As a partial, evaluation-only robustness check that does not require retraining, we re-evaluated the full model at an earlier retained checkpoint of the same training run. The orderings attributable to the two studied components persist at that checkpoint: the earlier full-model checkpoint still outperforms Ablation~A and Ablation~B on median NCC, the failure tail, and the phase-drift curves. The small aggregate margin over MTP-only does not clearly persist, however, at the earlier checkpoint the full model's median NCC ($0.950$) falls slightly below Ablation~C's ($0.954$), while its flip rate ($8.2\%$) and phase drift remain lower. The $+0.002$ Full-vs.-C margin should therefore be regarded as within checkpoint-to-checkpoint variability, whereas the head and coherence comparisons (Full vs.\ B and Full vs.\ A) are stable across checkpoints. This check probes sensitivity to the checkpoint-selection snapshot, not to the training seed. Fourth, our explanation of the coherence loss is empirical rather than a formal analysis of why magnitude coupling damps rollout drift. Fifth, the phase-drift failures identified above are only weakly constrained by a magnitude-only loss, motivating the phase-aware variants discussed above. These limitations bound the interpretation of the small component effects, but they do not undercut the main controlled finding that multi-token prediction is the dominant stabilizer within this autoregressive setup, and the remaining errors expose a concrete phase-aware direction for future objectives.

\section{Conclusion}
\label{sec:conclusion}
Through controlled ablations and paired test-set analysis, we find that multi-token prediction is the dominant stabilizer of long-horizon autoregressive rollout on oscillatory signals. Within the \textsc{SeismoGPT} training objective, the horizon-embedding hybrid head and cross-horizon STFT magnitude coherence loss each add a small increment to this effect, with modest additional complexity. The results should be read with their scope in mind: we have not shown that the component effects are large in absolute terms, that they hold across training seeds, datasets, or architectural classes, or that a magnitude-based coherence loss can correct the phase drift that dominates the lowest-NCC failures. The natural follow-up is, therefore, a phase-aware coherence objective evaluated across multiple seeds, real recordings, and external sequence-model baselines.



\bibliography{tmlr}

@inproceedings{bengio2015scheduled,
  title     = {Scheduled Sampling for Sequence Prediction with Recurrent Neural Networks},
  author    = {Bengio, Samy and Vinyals, Oriol and Jaitly, Navdeep and Shazeer, Noam},
  booktitle = {Advances in Neural Information Processing Systems (NeurIPS)},
  year      = {2015},
  eprint    = {1506.03099},
  archivePrefix = {arXiv},
  primaryClass  = {cs.LG}
}

@inproceedings{venkatraman2015dad,
  title     = {Improving Multi-Step Prediction of Learned Time Series Models},
  author    = {Venkatraman, Arun and Hebert, Martial and Bagnell, J. Andrew},
  booktitle = {Proceedings of the AAAI Conference on Artificial Intelligence},
  year      = {2015}
}

@inproceedings{lamb2016professor,
  title     = {Professor Forcing: A New Algorithm for Training Recurrent Networks},
  author    = {Lamb, Alex and Goyal, Anirudh and Zhang, Ying and Zhang, Saizheng and Courville, Aaron and Bengio, Yoshua},
  booktitle = {Advances in Neural Information Processing Systems (NeurIPS)},
  year      = {2016},
  eprint    = {1610.09038},
  archivePrefix = {arXiv},
  primaryClass  = {stat.ML}
}

@inproceedings{arora2022exposure,
  title     = {Why Exposure Bias Matters: An Imitation Learning Perspective of Error Accumulation in Language Generation},
  author    = {Arora, Kushal and El Asri, Layla and Bahuleyan, Hareesh and Cheung, Jackie Chi Kit},
  booktitle = {Findings of the Association for Computational Linguistics (ACL)},
  year      = {2022},
  eprint    = {2204.01171},
  archivePrefix = {arXiv},
  primaryClass  = {cs.CL}
}

@misc{deo2024lossdecomposition,
  title         = {Harnessing Loss Decomposition for Long-Horizon Wave Predictions via Deep Neural Networks},
  author        = {Deo, Indu Kant and Jaiman, Rajeev K.},
  year          = {2024},
  eprint        = {2412.02924},
  archivePrefix = {arXiv},
  primaryClass  = {cs.LG},
  note          = {NeurIPS 2024 Workshop on Machine Learning and the Physical Sciences}
}

@inproceedings{qi2020prophetnet,
  title     = {ProphetNet: Predicting Future N-gram for Sequence-to-Sequence Pre-training},
  author    = {Qi, Weizhen and Yan, Yu and Gong, Yeyun and Liu, Dayiheng and Duan, Nan and Chen, Jiusheng and Zhang, Ruofei and Zhou, Ming},
  booktitle = {Findings of the Association for Computational Linguistics (EMNLP)},
  year      = {2020},
  eprint    = {2001.04063},
  archivePrefix = {arXiv},
  primaryClass  = {cs.CL}
}

@misc{gloeckle2024multitoken,
  title         = {Better \& Faster Large Language Models via Multi-token Prediction},
  author        = {Gloeckle, Fabian and Idrissi, Badr Youbi and Rozi{\`e}re, Baptiste and Lopez-Paz, David and Synnaeve, Gabriel},
  year          = {2024},
  eprint        = {2404.19737},
  archivePrefix = {arXiv},
  primaryClass  = {cs.CL}
}

@misc{deepseek2024v3,
  title         = {DeepSeek-V3 Technical Report},
  author        = {{DeepSeek-AI}},
  year          = {2024},
  eprint        = {2412.19437},
  archivePrefix = {arXiv},
  primaryClass  = {cs.CL}
}

@inproceedings{nie2023patchtst,
  title     = {A Time Series is Worth 64 Words: Long-term Forecasting with Transformers},
  author    = {Nie, Yuqi and Nguyen, Nam H. and Sinthong, Phanwadee and Kalagnanam, Jayant},
  booktitle = {International Conference on Learning Representations (ICLR)},
  year      = {2023},
  eprint    = {2211.14730},
  archivePrefix = {arXiv},
  primaryClass  = {cs.LG}
}

@inproceedings{das2024timesfm,
  title     = {A Decoder-Only Foundation Model for Time-Series Forecasting},
  author    = {Das, Abhimanyu and Kong, Weihao and Sen, Rajat and Zhou, Yichen},
  booktitle = {Proceedings of the International Conference on Machine Learning (ICML)},
  year      = {2024},
  eprint    = {2310.10688},
  archivePrefix = {arXiv},
  primaryClass  = {cs.LG}
}

@article{ansari2024chronos,
  title   = {Chronos: Learning the Language of Time Series},
  author  = {Ansari, Abdul Fatir and Stella, Lorenzo and Turkmen, Caner and Zhang, Xiyuan and Mercado, Pedro and Shen, Huibin and Shchur, Oleksandr and Rangapuram, Syama Sundar and Pineda Arango, Sebastian and Kapoor, Shubham and Zschiegner, Jasper and Maddix, Danielle C. and Wang, Hao and Mahoney, Michael W. and Torkkola, Kari and Wilson, Andrew Gordon and Bohlke-Schneider, Michael and Wang, Yuyang},
  journal = {Transactions on Machine Learning Research (TMLR)},
  year    = {2024},
  eprint  = {2403.07815},
  archivePrefix = {arXiv},
  primaryClass  = {cs.LG}
}

@inproceedings{yamamoto2020parallelwavegan,
  title     = {Parallel WaveGAN: A Fast Waveform Generation Model Based on Generative Adversarial Networks with Multi-Resolution Spectrogram},
  author    = {Yamamoto, Ryuichi and Song, Eunwoo and Kim, Jae-Min},
  booktitle = {IEEE International Conference on Acoustics, Speech and Signal Processing (ICASSP)},
  year      = {2020},
  eprint    = {1910.11480},
  archivePrefix = {arXiv},
  primaryClass  = {eess.AS}
}

@inproceedings{engel2020ddsp,
  title     = {DDSP: Differentiable Digital Signal Processing},
  author    = {Engel, Jesse and Hantrakul, Lamtharn and Gu, Chenjie and Roberts, Adam},
  booktitle = {International Conference on Learning Representations (ICLR)},
  year      = {2020},
  eprint    = {2001.04643},
  archivePrefix = {arXiv},
  primaryClass  = {cs.LG}
}

@misc{esmail2026seismogpt,
  title         = {Data-Driven Forecasting of Three-Component Seismograms Using Transformer Architectures},
  author        = {Esmail, W. and Russell, S. and Klinge, J. and Kappes, A. and Thomas, C.},
  year          = {2026},
  eprint        = {2606.02912},
  archivePrefix = {arXiv},
  primaryClass  = {physics.geo-ph}
}

@article{BENTAIEB20127067,
    title = {A review and comparison of strategies for multi-step ahead time series forecasting based on the NN5 forecasting competition},
    journal = {Expert Systems with Applications},
    volume = {39},
    number = {8},
    pages = {7067-7083},
    year = {2012},
    issn = {0957-4174},
    doi = {https://doi.org/10.1016/j.eswa.2012.01.039},
    url = {https://www.sciencedirect.com/science/article/pii/S0957417412000528},
    author = {Souhaib {Ben Taieb} and Gianluca Bontempi and Amir F. Atiya and Antti Sorjamaa},
}

@article{lim2021tft,
  title   = {Temporal Fusion Transformers for Interpretable Multi-horizon Time Series Forecasting},
  author  = {Lim, Bryan and Ar{\i}k, Sercan {\"O}. and Loeff, Nicolas and Pfister, Tomas},
  journal = {International Journal of Forecasting},
  volume  = {37},
  number  = {4},
  pages   = {1748--1764},
  year    = {2021},
  eprint  = {1912.09363},
  archivePrefix = {arXiv},
  primaryClass  = {stat.ML}
}

@misc{rasul2023lagllama,
  title         = {Lag-Llama: Towards Foundation Models for Probabilistic Time Series Forecasting},
  author        = {Rasul, Kashif and Ashok, Arjun and Williams, Andrew Robert and Ghonia, Hena and Bhagwatkar, Rishika and Khorasani, Arian and Bayazi, Mohammad Javad Darvishi and Adamopoulos, George and Riachi, Roland and Hassen, Nadhir and Bilo{\v{s}}, Marin and Garg, Sahil and Schneider, Anderson and Chapados, Nicolas and Drouin, Alexandre and Zantedeschi, Valentina and Nevmyvaka, Yuriy and Rish, Irina},
  year          = {2023},
  eprint        = {2310.08278},
  archivePrefix = {arXiv},
  primaryClass  = {cs.LG}
}

@inproceedings{woo2024moirai,
  title     = {Unified Training of Universal Time Series Forecasting Transformers},
  author    = {Woo, Gerald and Liu, Chenghao and Kumar, Akshat and Xiong, Caiming and Savarese, Silvio and Sahoo, Doyen},
  booktitle = {Proceedings of the International Conference on Machine Learning (ICML)},
  year      = {2024},
  eprint    = {2402.02592},
  archivePrefix = {arXiv},
  primaryClass  = {cs.LG}
}

@misc{vlachas2023learning,
  title         = {Learning from Predictions: Fusing Training and Autoregressive Inference for Long-Term Spatiotemporal Forecasts},
  author        = {Vlachas, Pantelis R. and Koumoutsakos, Petros},
  year          = {2023},
  eprint        = {2302.11101},
  archivePrefix = {arXiv},
  primaryClass  = {cs.LG}
}

@article{zhu2019phasenet,
  title   = {PhaseNet: A Deep-Neural-Network-Based Seismic Arrival-Time Picking Method},
  author  = {Zhu, Weiqiang and Beroza, Gregory C.},
  journal = {Geophysical Journal International},
  volume  = {216},
  number  = {1},
  pages   = {261--273},
  year    = {2019},
  eprint  = {1803.03211},
  archivePrefix = {arXiv},
  primaryClass  = {physics.geo-ph}
}

@article{mousavi2020earthquake,
  title   = {Earthquake Transformer---An Attentive Deep-Learning Model for Simultaneous Earthquake Detection and Phase Picking},
  author  = {Mousavi, S. Mostafa and Ellsworth, William L. and Zhu, Weiqiang and Chuang, Lindsay Y. and Beroza, Gregory C.},
  journal = {Nature Communications},
  volume  = {11},
  number  = {1},
  pages   = {3952},
  year    = {2020}
}

@misc{liu2024seislm,
  title         = {SeisLM: a Foundation Model for Seismic Waveforms},
  author        = {Liu, Tianlin and M{\"u}nchmeyer, Jannes and Laurenti, Laura and Marone, Chris and de Hoop, Maarten V. and Dokmani{\'c}, Ivan},
  year          = {2024},
  eprint        = {2410.15765},
  archivePrefix = {arXiv},
  primaryClass  = {physics.geo-ph}
}

@inproceedings{gu2023mamba,
  title     = {Mamba: Linear-Time Sequence Modeling with Selective State Spaces},
  author    = {Gu, Albert and Dao, Tri},
  booktitle = {Conference on Language Modeling (COLM)},
  year      = {2024},
  eprint    = {2312.00752},
  archivePrefix = {arXiv},
  primaryClass  = {cs.LG}
}

@inproceedings{brandstetter2022mppde,
  title         = {Message Passing Neural PDE Solvers},
  author        = {Brandstetter, Johannes and Worrall, Daniel E. and Welling, Max},
  booktitle     = {International Conference on Learning Representations (ICLR)},
  year          = {2022},
  eprint        = {2202.03376},
  archivePrefix = {arXiv},
  primaryClass  = {cs.LG}
}

@inproceedings{ai2023antiwrapping,
  title         = {Neural Speech Phase Prediction Based on Parallel Estimation Architecture and Anti-Wrapping Losses},
  author        = {Ai, Yang and Ling, Zhen-Hua},
  booktitle     = {IEEE International Conference on Acoustics, Speech and Signal Processing (ICASSP)},
  year          = {2023},
  eprint        = {2211.15974},
  archivePrefix = {arXiv},
  primaryClass  = {eess.AS}
}

@article{Hendrycks:2016qxa,
    author = "Hendrycks, Dan and Gimpel, Kevin",
    title = "{Gaussian Error Linear Units (GELUs)}",
    eprint = "1606.08415",
    archivePrefix = "arXiv",
    primaryClass = "cs.LG",
    month = "6",
    year = "2016"
}

@article{su2024roformer,
  author  = {Su, Jianlin and Ahmed, Murtadha and Lu, Yu and Pan, Shengfeng and Bo, Wen and Liu, Yunfeng},
  title   = {RoFormer: Enhanced transformer with Rotary Position Embedding},
  journal = {Neurocomputing},
  year    = {2024},
  volume  = {568},
  pages   = {127063},
  doi     = {10.1016/j.neucom.2023.127063}
}

@inproceedings{loshchilov2019decoupled,
  author    = {Loshchilov, Ilya and Hutter, Frank},
  title     = {Decoupled Weight Decay Regularization},
  booktitle = {International Conference on Learning Representations (ICLR)},
  year      = {2019},
  url       = {https://openreview.net/forum?id=Bkg6RiCqY7}
}

@article{welch1967use,
  title={The use of fast {Fourier} transform for the estimation of power spectra: {A} method based on time averaging over short, modified periodograms},
  author={Welch, Peter},
  journal={IEEE Transactions on Audio and Electroacoustics},
  volume={15},
  number={2},
  pages={70--73},
  year={1967},
  publisher={IEEE},
  doi={10.1109/TAU.1967.1161901}
}

@article{ba2016layer,
  title   = {Layer Normalization},
  author  = {Ba, Jimmy Lei and Kiros, Jamie Ryan and Hinton, Geoffrey E},
  journal = {arXiv preprint arXiv:1607.06450},
  year    = {2016},
  url     = {https://arxiv.org/abs/1607.06450}
}
\bibliographystyle{tmlr}

\newpage

\appendix

\section{Training protocol and hyperparameters}
\label{app:hparams}
All configurations share the backbone, tokenization, optimizer, schedule, and data in
Table~\ref{tab:hparams}, and are trained to convergence under early stopping (patience 5, monitoring validation loss) on a shared, event-disjoint validation split. They differ only in the prediction head and the active loss terms, as defined in Section~\ref{sec:exp-setup}.

\begin{table}[h]
\centering
\caption{Shared hyperparameters across all configurations.}
\label{tab:hparams}
\begin{tabular}{ll}
\toprule
Backbone & causal transformer, RoPE, pre-norm \\
$d_{\text{model}}$ / heads / layers & $512$ / $8$ / $8$ \\
Feed-forward expansion / dropout & $4\times$ / $0.1$ \\
Token size $K$ / channels $C$ / context $T$ & $16$ / $3$ / $320$ \\
Horizons $H$ / horizon decay $\beta$ & $4$ / $0.6$ \\
STFT FFT sizes / $\lambda_{\text{stft}}$ / $\lambda_{\Delta}$ / $\lambda_{\text{coh}}$
  & $\{128,256,512\}$ / $0.05$ / $0.05$ / $0.1$ \\
Optimiser / peak LR & AdamW / $10^{-4}$ \\
Schedule & cosine warm restarts ($T_0{=}10$, $T_{\text{mult}}{=}2$), $1000$-step warmup \\
Gradient clip / precision / batch size & $1.0$ / bf16-mixed / $32$ \\
\bottomrule
\end{tabular}
\end{table}

\section{Per-horizon analysis}
\label{app:horizon}
We computed each metric separately at every prediction horizon $h \in \{0,\dots,3\}$. The configurations are essentially indistinguishable at every horizon (maximum pairwise gap $<0.001$ NCC), confirming that the contributions act on rollout dynamics rather than per-step accuracy. We therefore omit a per-horizon column from Table~\ref{tab:ablation}.

\section{Worst-case failures}
\label{app:failures}

\begin{figure}[h]
    \centering
    \includegraphics[width=0.8\linewidth]{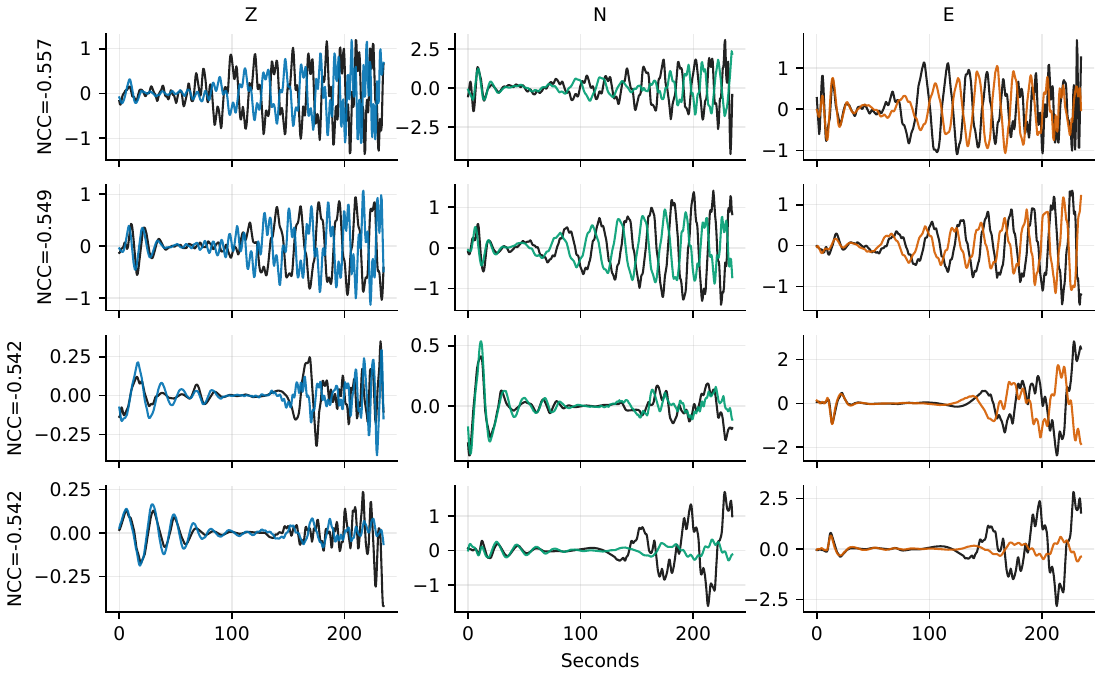}
    \caption{Worst-case rollouts (lowest NCC), prediction versus ground truth, all three components. NCC is negative (e.g.\ $-0.557$): late high-energy phases become polarity-inverted, which a magnitude-based spectral loss cannot penalize by construction (Section~\ref{sec:discussion}).}
    \label{fig:failure}
\end{figure}


\end{document}